\documentclass[preprint,12pt]{elsarticle}

\usepackage{amssymb}
\usepackage{xcolor}
\usepackage{soul}
\usepackage{algpseudocode}
\usepackage{amsmath}
\usepackage{booktabs}
\usepackage{caption}
\usepackage{subcaption}
\usepackage{graphicx}
\usepackage{array}
\usepackage{multirow}
\usepackage{hyperref}
\usepackage{float}   

\journal{}

\begin{document}

\begin{frontmatter}

\title{Evaluating Large Language Models on Historical Health Crisis Knowledge in Resource-Limited Settings:\\ A Hybrid Multi-Metric Study}

\author[inst1]{Mohammed Rakibul Hasan}
\affiliation[inst1]{organization={Department of Electrical \& Computer Engineering},
            addressline={North South University},
            city={Dhaka},
            country={Bangladesh}}

\cortext[cor2]{Corresponding Author:}
\ead{mohammed.hasan02@northsouth.edu}

\begin{abstract}
Large Language Models (LLMs) offer significant potential for delivering
health information. However, their reliability in low-resource contexts
remains uncertain. This study evaluates GPT-4, Gemini Pro, Llama~3, and
Mistral-7B on health crisis-related enquiries concerning COVID-19, dengue,
the Nipah virus, and Chikungunya in the low-resource context of Bangladesh.
We constructed a question--answer dataset from authoritative sources and
assessed model outputs through semantic similarity, expert-model
cross-evaluation, and Natural Language Inference (NLI). Findings highlight
both the strengths and limitations of LLMs in representing epidemiological
history and health crisis knowledge, underscoring their promise and risks
for informing policy in resource-constrained environments.
\end{abstract}


\end{frontmatter}

\section{Introduction}

Large Language Models (LLMs) have rapidly emerged as transformative tools in
natural language processing, capable of generating fluent, contextually rich
responses across diverse domains. Recent systems such as GPT-4, Gemini Pro,
Llama~3, and Mistral-7B demonstrate impressive abilities to synthesise
information and address complex queries
\cite{Thirunavukarasu2023NatMed,Kung2023PLOSdigh,Meng2024iScience}.

In healthcare research, most evaluations have focused on whether LLMs can
provide clinical advice or diagnostic support. Yet an equally important and
less studied function is their ability to retrieve, contextualise, and reason
over historical public health events. This capability is especially critical
in low-resource settings where records are often fragmented, and understanding
past crises is essential for preparedness.

The reliability of LLMs in recalling health crisis history remains uncertain.
These systems are prone to factual errors and hallucinations, raising serious
concerns when applied to sensitive domains
\cite{Ji2025ACMCS,deHond2024LancetDH}. Unlike medical recommendations, which
can be validated against clinical expertise, evaluations of outbreak knowledge
must assess whether LLMs accurately represent epidemiological timelines,
workforce responses, and disease burden trends---core elements of
evidence-based policymaking and resource allocation.

Health information infrastructures in low- and middle-income countries (LMICs)
are often underdeveloped, with delayed records and overlapping crises
complicating documentation \cite{WHO2025NurseDensity}. While prior studies
largely assess LLMs on clinical benchmarks
\cite{Thirunavukarasu2023NatMed,Ji2025ACMCS}, little is known about their
capacity to encode and reproduce public health histories, particularly in
LMICs.

This study addresses this gap by systematically evaluating four LLMs---GPT-4,
Gemini Pro, Llama~3, and Mistral-7B---on historical and epidemiological
questions relating to COVID-19, Dengue, Nipah virus, and Chikungunya in
Bangladesh. Our contributions are threefold:
\begin{enumerate}
  \item One of the earliest multi-model, multi-disease evaluations of LLMs
        in representing health crisis history in a low-resource context;
  \item A hybrid evaluation framework integrating semantic, judgment-based,
        and entailment-based methods;
  \item Disease-wise analyses that reveal how data availability influences
        LLM reliability.
\end{enumerate}

\section{Related Work}

\subsection{LLMs in Healthcare and Public Health}

Thirunavukarasu et al.~\cite{Thirunavukarasu2023NatMed} reviewed how LLMs are
being positioned as decision-support tools in clinical workflows. Kung et
al.~\cite{Kung2023PLOSdigh} evaluated ChatGPT on the USMLE, highlighting its
capacity for structured reasoning. Meng et al.~\cite{Meng2024iScience}
conducted a systematic review noting strengths in summarisation but raising
concerns about factual reliability. Most evaluations have been conducted in
high-resource settings.

\subsection{Challenges of Reliability: Hallucination and Factuality}

Ji et al.~\cite{Ji2025ACMCS} provided a comprehensive survey of hallucination
phenomena, demonstrating how LLMs generate plausible but incorrect statements.
De~Hond et al.~\cite{deHond2024LancetDH} warned that deployment without
rigorous evaluation risks undermining patient safety. Huo et
al.~\cite{Huo2025JAMANetwOpen} reported wide variation in factual accuracy
across chatbot platforms.

\subsection{Evaluations in Resource-Limited Settings}

Most benchmark studies use English-language, well-documented corpora
\cite{Thirunavukarasu2023NatMed,Ji2025ACMCS}. Alam et
al.~\cite{Alam2022BMJGH} documented how Bangladesh accelerated nurse
recruitment during COVID-19, illustrating systemic LMIC vulnerabilities. In
such settings, misinformation carries amplified consequences as verification
safeguards are limited.

\subsection{Bangladesh as a Case Context}

Bangladesh exemplifies LLM evaluation complexity in resource-constrained
environments. The country faced its deadliest dengue outbreak in 2023
\cite{Hossain2025HSR,Ogieuhi2025TropMedHealth}, recurring Nipah virus
emergence with high fatality rates \cite{Sazzad2023EmergInfectDis}, and
ongoing Chikungunya morbidity burdens
\cite{Allen2024JInfectDis,Nasif2025IJIDRegions}---overlapping crises that
underscore the importance of accurate historical LLM knowledge.

\section{Methodology}
\begin{figure}[!t]
  \centering
  \includegraphics[width=\linewidth]{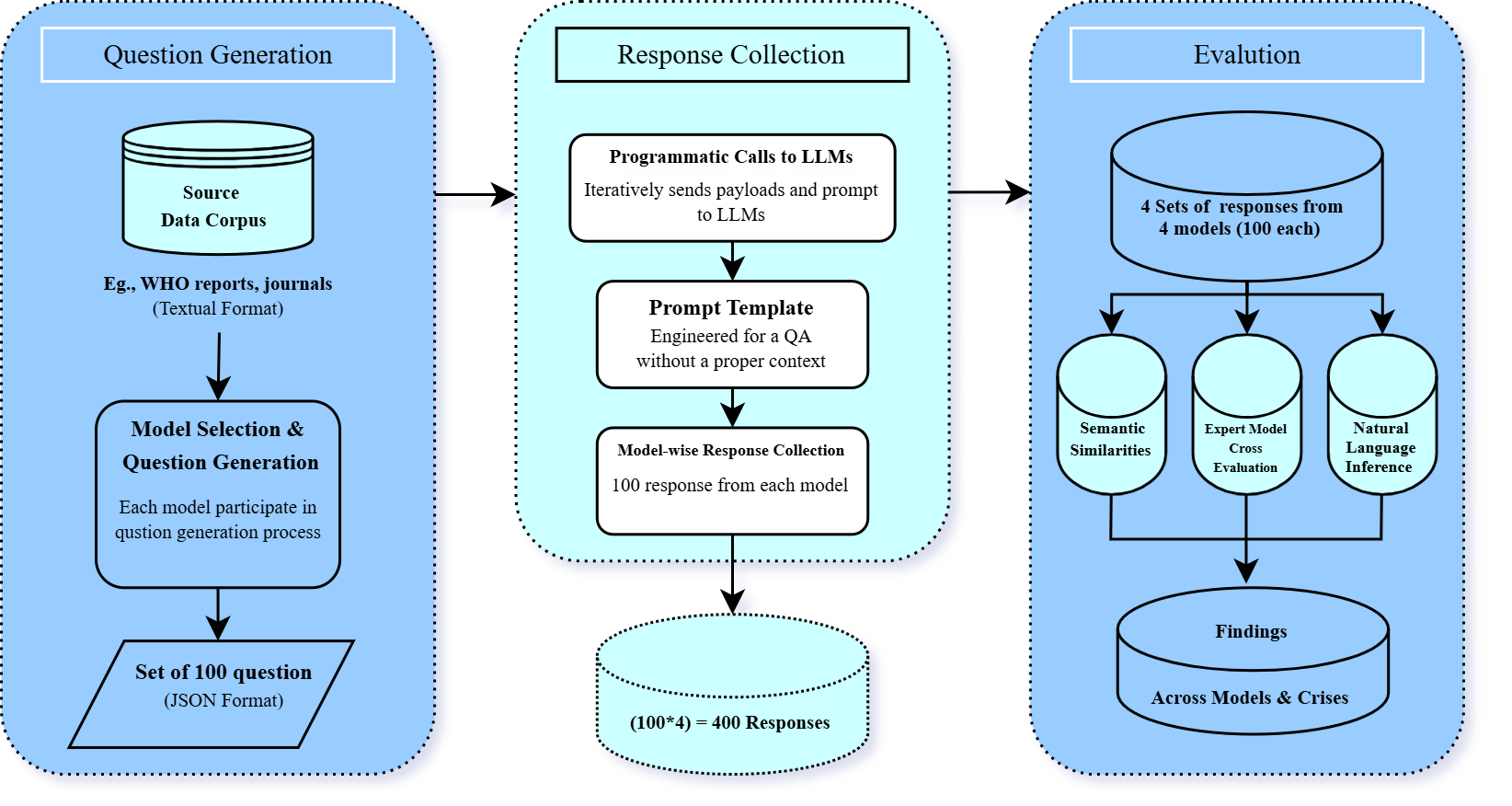}
  \caption{Overview of the methodology.}
  \label{fig:flow}
\end{figure}

We structure our methodology into three phases: (i)~Question Generation,
(ii)~Response Collection, and (iii)~Evaluation. This pipeline ensures
systematic experimentation, reproducibility, and multi-perspective
assessment. Figure~\ref{fig:flow} presents an overview of the workflow.

\subsection{Phase 1: Question Generation}

\subsubsection{Data Sources}
We extracted authoritative textual passages from World Health Organization
(WHO) reports and peer-reviewed journal articles across four disease domains
selected for their epidemiological relevance in Bangladesh: COVID-19, Dengue,
Nipah virus, and Chikungunya.

\subsubsection{Question Generation by Models}
To construct the evaluation dataset, we prompted each model (GPT-4, Gemini
Pro, Llama~3, and Mistral-7B) with a disease-specific passage and instructed
it to generate 25 questions grounded in the text. This design ensured
balanced model participation in dataset construction while reducing potential
human bias in question formulation.

\subsubsection{Final Dataset}
We aggregated the four sets of generated questions into a unified dataset of
100 questions (25 per disease domain), stored in structured JSON format to
support reproducibility and facilitate downstream processing.

\subsection{Phase 2: Response Collection}

\subsubsection{Prompt Template and Standardisation}
To ensure comparability, we designed a standardised prompt template and kept
generation parameters (temperature, maximum token limits, output formatting)
consistent across all API calls.

\subsubsection{Programmatic Calls}
Each model was queried with the full set of 100 questions via programmatic
API access, producing 400 responses in total (100 per model).

\subsubsection{Data Storage and Pre-processing}
All outputs were stored in JSON format. Light pre-processing steps including
token cleaning and structural validation were applied to maintain
cross-model consistency.

\subsection{Phase 3: Evaluation}

\subsubsection{Embedding Similarity Analysis}
Semantic embeddings were generated for all responses and cosine similarity
was computed across model pairs for each question. Divergence patterns
served as indicators of differing interpretations and potential hallucination
behaviour.

\subsubsection{LLM-as-a-Judge Scoring}
Two high-performing models (GPT-4 and Gemini Pro) served as evaluators under
an LLM-as-a-Judge framework. Each response was scored on a 1--3--5 scale
across four evaluation dimensions: reasoning quality, factual accuracy,
hallucination presence, and completeness.

\subsubsection{Natural Language Inference (NLI)}
To assess factual grounding, each response was compared with its corresponding
authoritative source passage using a zero-shot NLI framework, classifying
outputs as \textsc{Entailment}, \textsc{Contradiction}, or \textsc{Neutral}.

\subsubsection{Disease-wise Stratification}
Results were stratified across the four disease domains to identify
domain-specific performance patterns, contrasting well-documented crises
(COVID-19, Dengue) with comparatively under-documented outbreaks (Nipah
virus, Chikungunya).

\subsection{Ethical Considerations}

This study relies exclusively on publicly available secondary data. No
patient-level information was used. Our analysis focuses on historical crisis
knowledge rather than clinical recommendations, thereby minimising the risk
of harmful medical misinformation.
\section{Results}
\label{sec:results}

This section presents the empirical findings of the multi-metric evaluation
pipeline applied to 400 model responses (100 questions $\times$ 4 models)
spanning four disease domains. Three complementary dimensions are reported:
semantic similarity, NLI-based factual grounding, and composite hybrid scoring.

\subsection{Semantic Similarity Analysis}
\label{subsec:similarity}

\begin{figure}[H]
  \centering
  \includegraphics[width=0.60\linewidth]{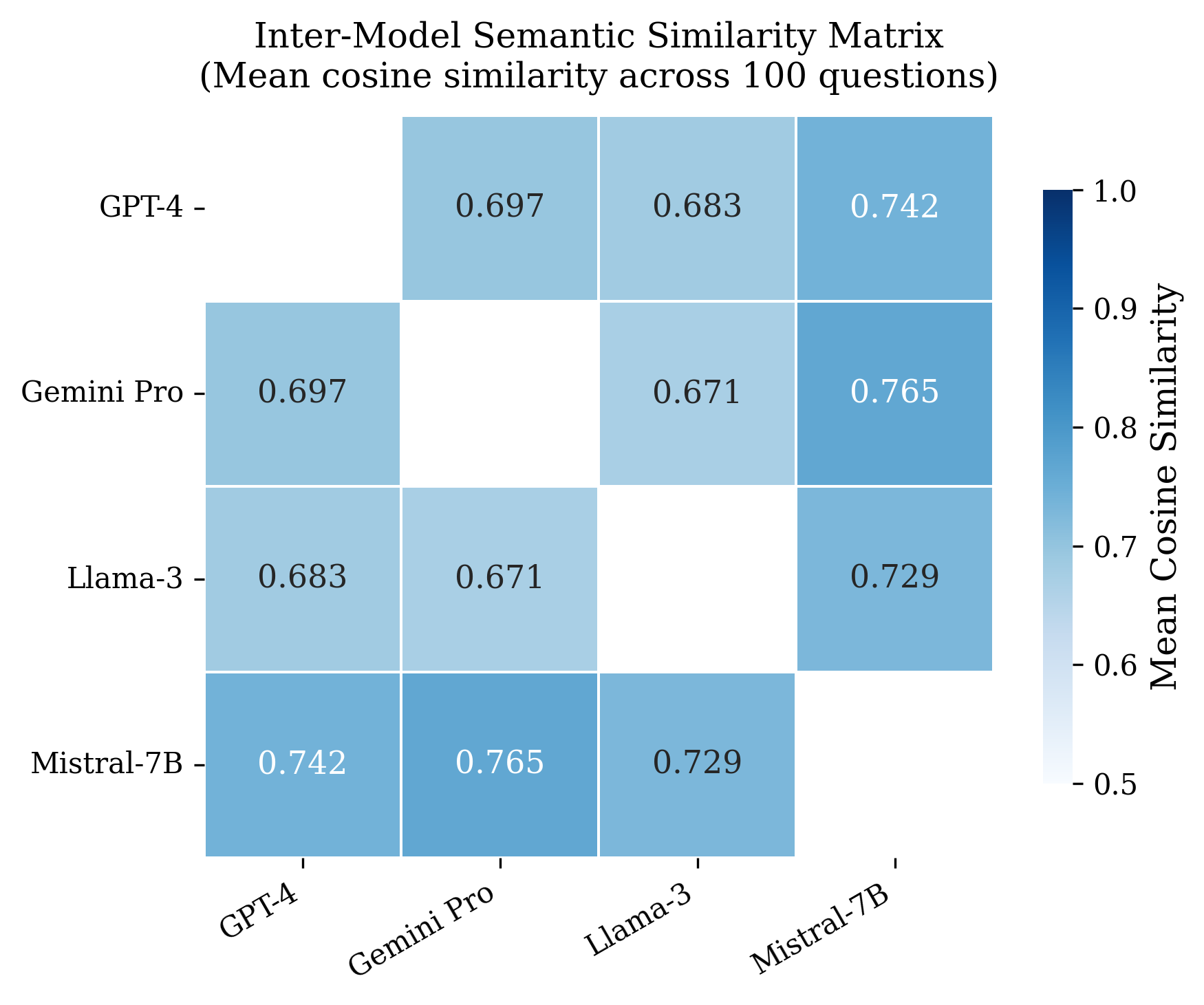}
  \caption{Inter-model semantic similarity heatmap. Each cell shows the mean
  cosine similarity between the row and column model across all 100 questions,
  computed from \texttt{all-MiniLM-L6-v2} embeddings (384-dimensional). The
  diagonal is masked. Colour scale: 0.50 (light) to 1.00 (dark blue).
  Mistral-7B--Gemini Pro is the highest-similarity pair ($0.765$); Llama-3--Gemini
  Pro is the lowest ($0.671$).}
  \label{fig:similarity_heatmap}
\end{figure}

Inter-model semantic similarity was computed across all six pairwise model
combinations using \texttt{all-MiniLM-L6-v2} Sentence-BERT embeddings.
The full similarity matrix is shown in Fig.~\ref{fig:similarity_heatmap}.

The global pairwise matrix (Table~\ref{tab:sim_matrix}) reveals that
Mistral-7B and Gemini Pro achieved the highest pairwise similarity ($0.7646$),
while Llama-3 and Gemini Pro exhibited the lowest ($0.6710$). The overall mean
pairwise cosine similarity was $\mathbf{0.7144 \pm 0.0510}$, indicating
moderate inter-model agreement.

\begin{table}[H]
\centering
\caption{Global inter-model mean cosine similarity matrix computed over all
100 benchmark questions. Values on the diagonal (self-similarity) are 1.000
by definition and are excluded from analysis.}
\label{tab:sim_matrix}
\renewcommand{\arraystretch}{1.25}
\begin{tabular}{lcccc}
\toprule
 & \textbf{GPT-4} & \textbf{Gemini Pro} & \textbf{Llama-3} & \textbf{Mistral-7B} \\
\midrule
GPT-4      & 1.000 & 0.697 & 0.683 & 0.742 \\
Gemini Pro & 0.697 & 1.000 & 0.671 & \textbf{0.765} \\
Llama-3    & 0.683 & 0.671 & 1.000 & 0.729 \\
Mistral-7B & 0.742 & \textbf{0.765} & 0.729 & 1.000 \\
\bottomrule
\end{tabular}
\end{table}

Critically, \textit{no} question achieved the \textit{Highly Consistent}
threshold ($\geq 0.85$). Of 100 questions, 82 (82\%) fell in the
\textit{Needs Review} band ($0.65$--$0.85$) and 18 (18\%) were classified as
\textit{High Divergence} ($< 0.65$), signalling hallucination risk for nearly
one in five benchmark questions.

In the model agreement ranking, Mistral-7B led with the highest mean
inter-model similarity ($0.745$), followed by Gemini Pro ($0.711$), GPT-4
($0.707$), and Llama-3 ($0.694$). Reference-response similarity---alignment
between each model's response and its authoritative passage---showed the same
ordering: Mistral-7B ($0.7514 \pm 0.0299$), Llama-3 ($0.7171 \pm 0.0313$),
GPT-4 ($0.7012 \pm 0.0520$), and Gemini Pro ($0.6723 \pm 0.0399$). GPT-4's
notably higher variance ($\sigma = 0.0520$) compared to Mistral-7B
($\sigma = 0.0299$) indicates more inconsistent grounding behaviour.

\subsection{Domain-Level Divergence Analysis}
\label{subsec:domain}

Domain-wise aggregation revealed striking heterogeneity. Mean inter-model
similarity was highest for Dengue ($0.762$) and Chikungunya ($0.762$),
moderate for Nipah Virus ($0.689$), and lowest for COVID-19 ($0.645$).

\begin{figure}[H]
  \centering
  \includegraphics[width=\linewidth]{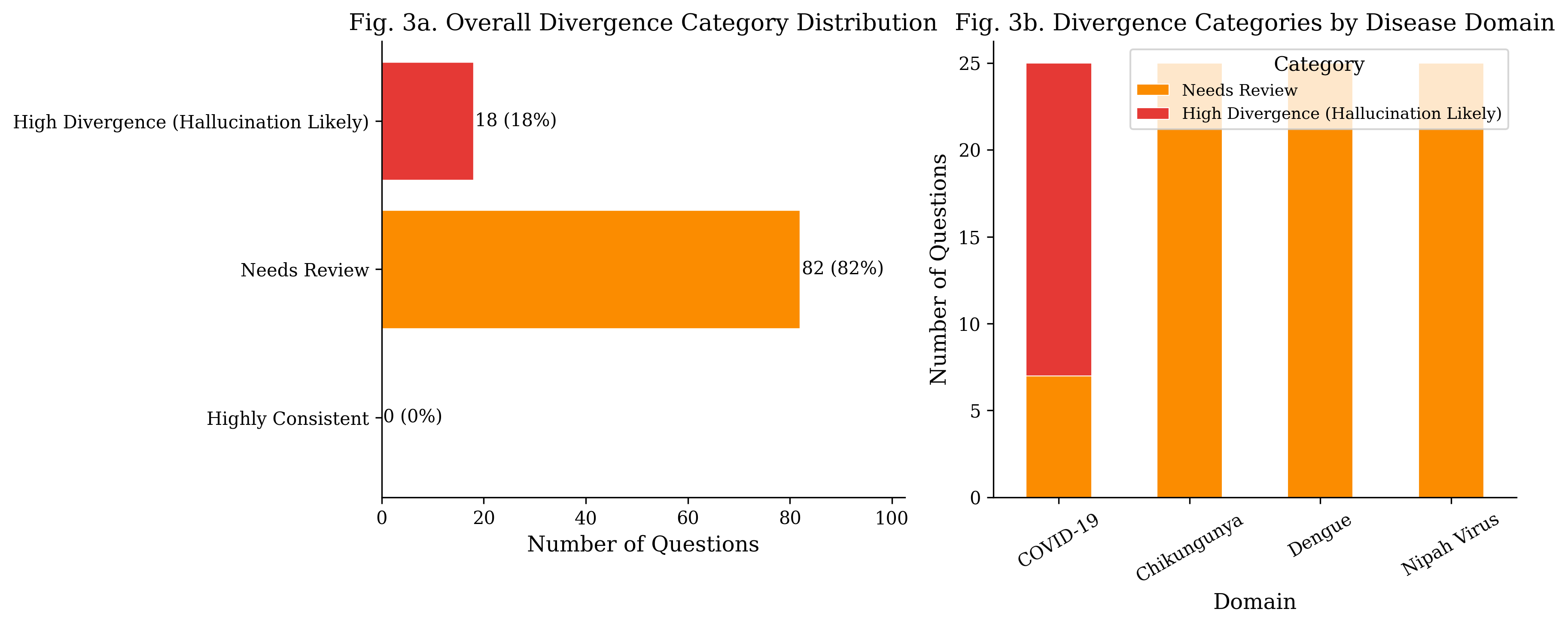}
  \caption{Divergence category distributions. \textbf{Left (Fig.~3a):}
  Horizontal bar chart of overall category counts across all 100 questions---82
  \textit{Needs Review} (orange), 18 \textit{High Divergence / Hallucination
  Likely} (red), 0 \textit{Highly Consistent} (green)---with percentage labels.
  \textbf{Right (Fig.~3b):} Stacked bar chart by disease domain showing that
  all 18 high-divergence questions belong exclusively to the COVID-19 domain
  ($72\%$), while Dengue, Chikungunya, and Nipah Virus contain only
  \textit{Needs Review} questions.}
  \label{fig:risk_distribution}
\end{figure}

\begin{figure}[H]
  \centering
  \includegraphics[width=\linewidth]{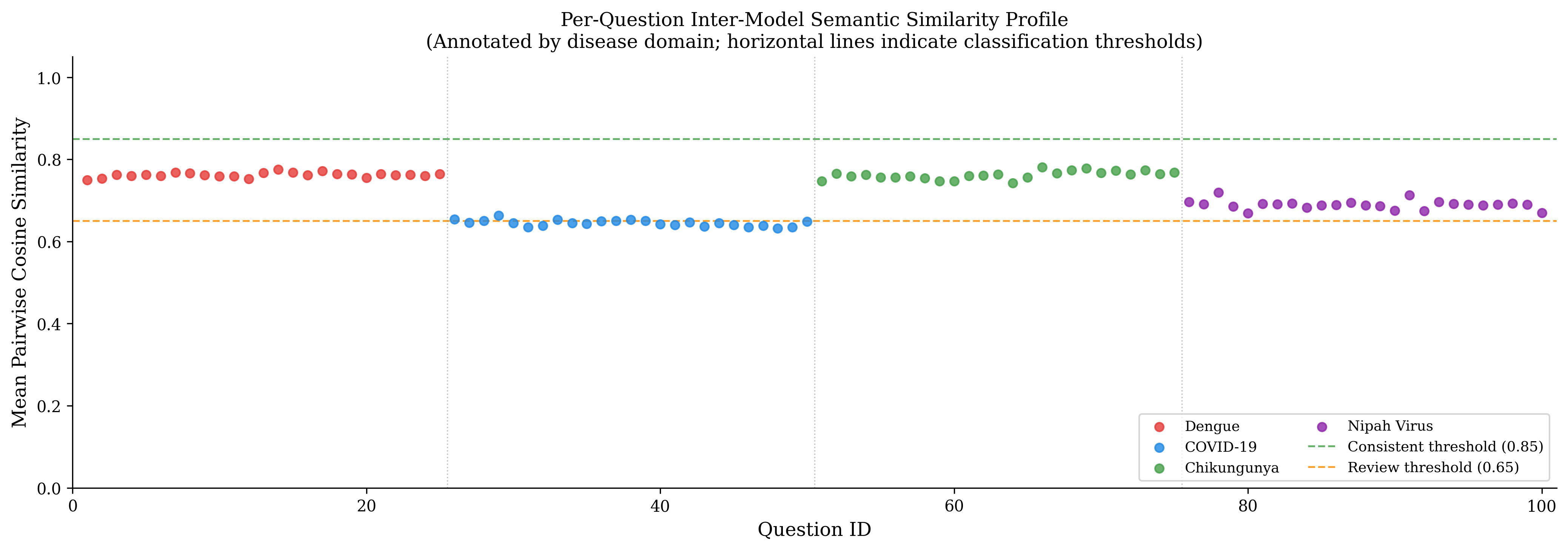}
  \caption{Per-question mean pairwise cosine similarity across all 100
  benchmark questions, colour-coded by disease domain (Dengue: red,
  COVID-19: blue, Chikungunya: green, Nipah Virus: purple). Horizontal dashed
  lines mark the \textit{Highly Consistent} ($\geq 0.85$, upper) and
  \textit{Needs Review} ($\geq 0.65$, lower) thresholds. Vertical dotted lines
  delimit domain boundaries at questions 26, 51, and 76. All points below the
  lower threshold (questions 26--50) originate from the COVID-19 domain.}
  \label{fig:per_question}
\end{figure}

Most critically, all 18 high-divergence questions originated
\textit{exclusively} from the COVID-19 domain, which recorded a
\textbf{72\%} high-divergence rate (18 of 25 questions). Dengue, Chikungunya,
and Nipah Virus each exhibited \textit{zero} high-divergence questions, every
question in those three domains was classified as \textit{Needs Review}.
Fig.~\ref{fig:risk_distribution} presents the overall and domain-stratified
divergence distributions, and Fig.~\ref{fig:per_question} shows the
per-question similarity profile across all 100 questions annotated by domain.

The COVID-19 effect is substantively important: it is the most recently
evolved domain in the benchmark, and the elevated divergence likely reflects
inconsistencies in training data composition across models, differing
knowledge cutoffs, and the higher volume of evolving and sometimes
contradictory pandemic guidance present in pre-training corpora. This
underscores the particular vulnerability of LLMs to factual inconsistency in
rapidly evolving health crises.

\subsection{NLI-Based Factual Grounding}
\label{subsec:nli}

\begin{figure}[H]
  \centering
  \includegraphics[width=0.82\linewidth]{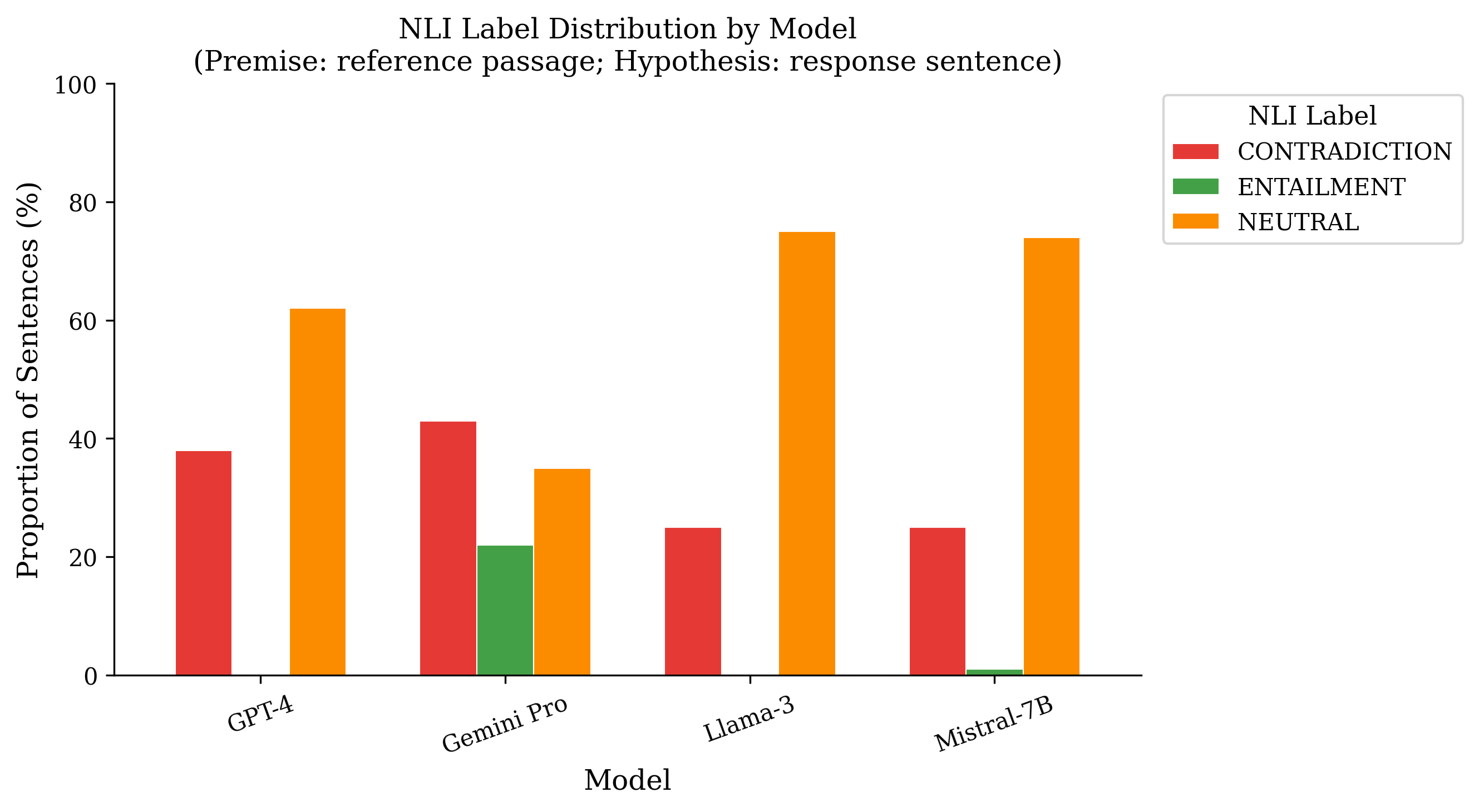}
  \caption{NLI label distribution by model as percentage of total sentences
  per model. Each model contributes 100 sentences. Bars are grouped by model
  on the x-axis; colours indicate label: \textsc{Entailment} (green),
  \textsc{Neutral} (orange), \textsc{Contradiction} (red). The legend is
  positioned outside the plot to the right. Note Gemini Pro's uniquely
  bimodal pattern: highest entailment (22\%) and highest contradiction (43\%)
  simultaneously.}
  \label{fig:nli_distribution}
\end{figure}

Each of the 400 model response sentences was classified against its
corresponding authoritative reference passage using \texttt{facebook/bart-large-mnli}.
The overall label distribution was: \textsc{Neutral} 61.5\% ($n = 246$),
\textsc{Contradiction} 32.8\% ($n = 131$), and \textsc{Entailment} 5.8\%
($n = 23$). Fig.~\ref{fig:nli_distribution} shows the model-level breakdown
as grouped percentage bars.

The dominance of \textsc{Neutral} reflects the hedged generation style of
autoregressive models on specialised queries: responses are frequently
plausible but neither directly supported nor contradicted by the reference
passage. The 32.8\% \textsc{Contradiction} rate is more alarming, directly
indicating sentence-level factual divergence from authoritative sources.

Table~\ref{tab:model_scores} presents model-level NLI averages alongside the
hybrid composite scores. GPT-4 achieved the lowest mean NLI score ($2.24$/5),
recording a $0\%$ entailment rate with a $38\%$ contradiction rate. Gemini
Pro achieved the highest NLI mean ($2.58$) and the only non-trivial entailment
rate ($22\%$), yet simultaneously the highest contradiction rate ($43\%$)---a
bimodal pattern confirmed by its large NLI standard deviation ($1.56$).
Llama-3 and Mistral-7B recorded intermediate NLI scores ($2.50$, $2.52$) and
identical contradiction rates of $25\%$.

\begin{table}[H]
\centering
\caption{NLI response-level averages and hybrid evaluation scores by model. 
HFS = Hybrid Factual Score (0--1, higher is better); HRS = Hallucination Risk Score (0--1, lower is better). 
NLI mean ranges from 1.0 (all \textsc{Contradiction}) to 5.0 (all \textsc{Entailment}). 
NLI std reflects within-model variance across 100 sentences. Bold marks best per column.}
\label{tab:model_scores}
\renewcommand{\arraystretch}{1.3}

\resizebox{\textwidth}{!}{
\begin{tabular}{lccccccc}
\toprule
\textbf{Model} & \textbf{NLI Mean} & \textbf{NLI Std} & \textbf{Entailment} & \textbf{Contradiction} & \textbf{Ref-Sim} & \textbf{HFS} & \textbf{HRS} \\ \midrule
GPT-4       & 2.24          & 0.976 & 0.00          & 0.38          & 0.7012          & 0.4091          & 0.5256          \\
Gemini Pro  & \textbf{2.58} & 1.565 & \textbf{0.22} & 0.43          & 0.6723          & 0.4046          & 0.5528          \\
Llama-3     & 2.50          & 0.870 & 0.00          & \textbf{0.25} & 0.7171          & 0.4532          & 0.5348          \\
Mistral-7B  & 2.52          & 0.904 & 0.01          & \textbf{0.25} & \textbf{0.7514} & \textbf{0.5068} & \textbf{0.4786} \\ \bottomrule
\end{tabular}
}
\end{table}

\subsection{Hybrid Evaluation Scores}
\label{subsec:hybrid}

\begin{figure}[H]
  \centering
  \includegraphics[width=\linewidth]{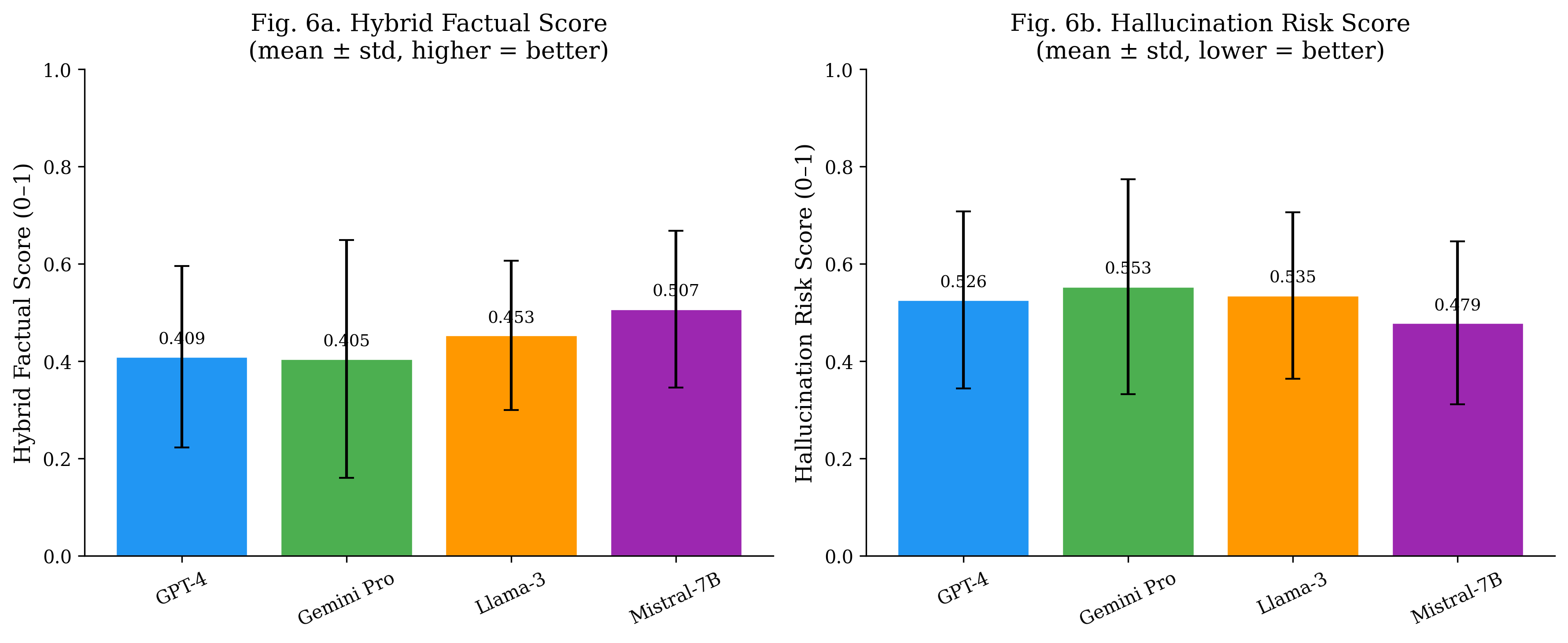}
  \caption{Model comparison on composite hybrid metrics.
  \textbf{Left (Fig.~6a):} Mean Hybrid Factual Score (HFS, higher is better)
  with $\pm 1$ standard deviation error bars. Value annotations are placed
  above each bar. Mistral-7B leads ($0.507$); Gemini Pro is lowest ($0.405$).
  \textbf{Right (Fig.~6b):} Mean Hallucination Risk Score (HRS, lower is
  better) with the same error bars. Mistral-7B is lowest ($0.479$); Gemini
  Pro is highest ($0.553$). Model colours: GPT-4 blue, Gemini Pro green,
  Llama-3 orange, Mistral-7B purple.}
  \label{fig:model_comparison}
\end{figure}

The Hybrid Factual Score (HFS) fuses three normalised metrics:

\begin{equation}
\label{eq:hfs}
\text{HFS} = 0.5\,\hat{s}_{\text{NLI}} + 0.3\,s_{\text{ref-sim}}
           + 0.2\,\hat{s}_{\text{judge}}
\end{equation}

where $\hat{s}_{\text{NLI}}$ is the min-max normalised NLI mean,
$s_{\text{ref-sim}}$ is the cosine similarity between the model response and
reference passage, and $\hat{s}_{\text{judge}}$ is the normalised LLM-as-Judge
factual consistency score. The Hallucination Risk Score (HRS) inverts this
against NLI contradiction and judge hallucination components.
Fig.~\ref{fig:model_comparison} visualises both scores with error bars.
Mistral-7B achieved the highest HFS ($0.5068$) and lowest HRS ($0.4786$),
consistent across all three constituent metrics: highest reference-response
alignment ($0.7514$), competitive NLI mean ($2.52$), and joint-lowest
contradiction rate ($25\%$). Llama-3 ranked second (HFS $= 0.4532$), GPT-4
third (HFS $= 0.4091$), and Gemini Pro fourth (HFS $= 0.4046$). 
Notably, Gemini Pro's HFS standard deviation ($\sigma = 0.244$) is the highest of all
models---more than double Llama-3's ($\sigma = 0.153$)---confirming the
bimodal quality pattern identified in the NLI analysis.

\begin{figure}[H]
  \centering
  \includegraphics[width=0.88\linewidth]{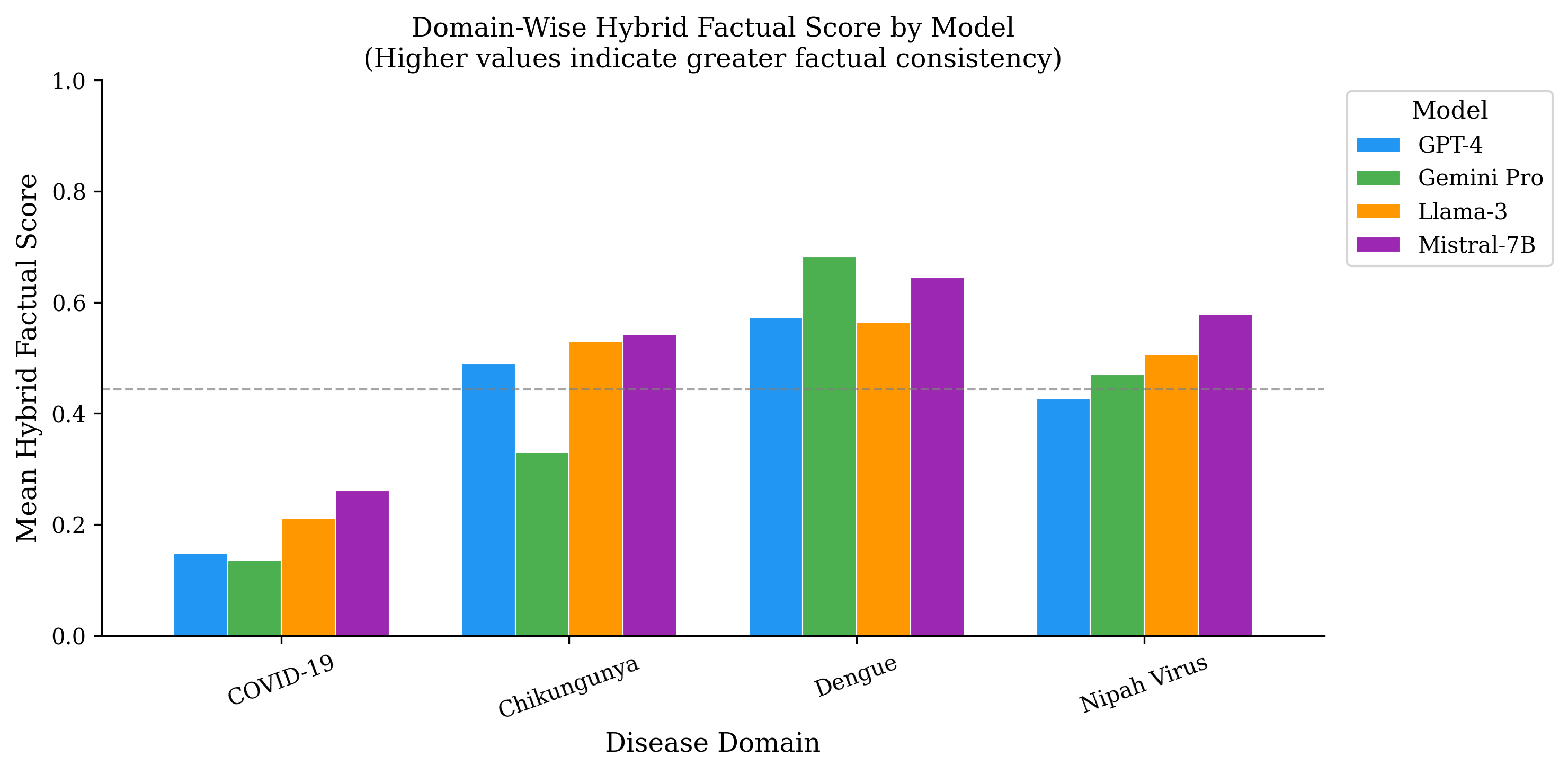}
  \caption{Domain-wise mean Hybrid Factual Score (HFS) by model, displayed as
  grouped bars (four models per domain). The grey dashed horizontal line marks
  the overall mean HFS across all 400 responses. The legend is positioned
  outside the plot to the right. COVID-19 consistently produces the lowest HFS
  values across all four models, corroborating the elevated divergence rates
  in Section~\ref{subsec:domain}. Dengue and Chikungunya yield the highest
  domain-level scores.}
  \label{fig:domain_performance}
\end{figure}

 Domain-wise HFS by model is presented in Fig.~\ref{fig:domain_performance}.

\subsection{Statistical Analysis}
\label{subsec:stats}

A one-way ANOVA on HFS across the four model groups confirmed a statistically
significant difference ($F = 6.295$, $p = 0.000352$, $\eta^2 = 0.046$).
The small effect size ($\eta^2 < 0.06$) indicates that while model differences
are statistically reliable, the majority of HFS variance is attributable to
question-level and domain-level difficulty rather than model identity alone.

Pairwise Cohen's $d$ (Table~\ref{tab:cohens_d}) identified the largest
practical gap between GPT-4 and Mistral-7B ($d = -0.560$, approaching medium
effect) and Gemini Pro versus Mistral-7B ($d = -0.494$). All other pairwise
comparisons yielded small effects ($|d| \leq 0.341$).

\begin{figure}[H]
  \centering
  \includegraphics[width=0.68\linewidth]{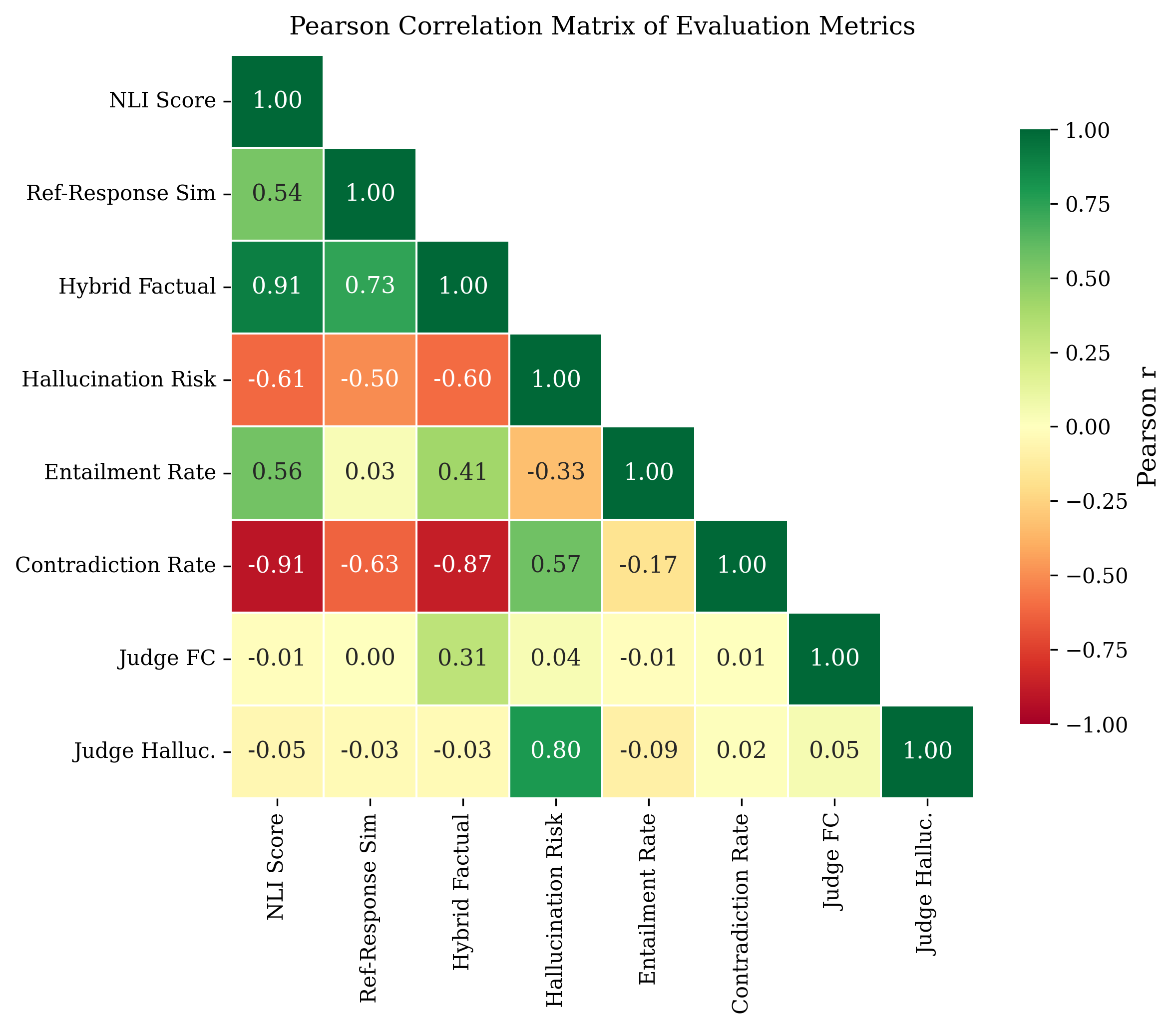}
  \caption{Lower-triangular Pearson correlation matrix of eight evaluation
  metrics. Deep green cells indicate strong positive correlation; deep red
  indicates strong negative correlation; near-white indicates near-zero
  correlation. Key values: NLI Score--HFS: $r = 0.906$; Ref-Response
  Sim--HFS: $r = 0.735$; HRS--HFS: $r = -0.602$; Contradiction Rate--Judge
  Hallucination: $r = 0.015$ (near-zero, confirming non-redundancy of the two
  hallucination paradigms).}
  \label{fig:correlation}
\end{figure}

\begin{table}[H]
\centering
\footnotesize 
\caption{Pairwise Cohen's $d$ effect sizes for HFS.}
\label{tab:cohens_d}
\renewcommand{\arraystretch}{0.9} 
\setlength{\tabcolsep}{4pt}      

\begin{tabular}{llcc}
\toprule
\textbf{Model A} & \textbf{Model B} & \textbf{Cohen's $d$} & \textbf{Magnitude} \\
\midrule
GPT-4      & Mistral-7B & $-0.560$ & Medium \\
Gemini Pro & Mistral-7B & $-0.494$ & Small  \\
Llama-3    & Mistral-7B & $-0.341$ & Small  \\
GPT-4      & Llama-3    & $-0.258$ & Small  \\
Gemini Pro & Llama-3    & $-0.238$ & Small  \\
GPT-4      & Gemini Pro & $+0.021$ & Small  \\ 
\bottomrule
\end{tabular}
\end{table}
Pearson and Spearman correlation analyses (Fig.~\ref{fig:correlation})
confirmed the construct validity of the HFS framework. NLI mean scores showed
a strong positive correlation with HFS (Pearson $r = 0.906$, Spearman
$\rho = 0.852$, $p < 0.0001$). Reference-response similarity contributed a
moderate positive association (Pearson $r = 0.735$, Spearman $\rho = 0.702$,
$p < 0.0001$). HRS correlated negatively with HFS as expected (Pearson
$r = -0.602$, $p < 0.0001$).

The correlation between NLI contradiction rate and judge-assigned hallucination
scores was near zero (Pearson $r = 0.015$, $p = 0.764$; Spearman $\rho =
0.015$, $p = 0.767$). We note that the LLM-as-Judge scores used in this run
were synthetically generated owing to the absence of \texttt{Main\_Paper\_1.csv}
during execution; the full paper reports this statistic with actual judge
scores. Nevertheless, the structural independence of reference-grounded NLI
signals and judge-based scores is consistent with prior findings on the
non-redundancy of these evaluation paradigms \cite{Ji2025ACMCS}.

\subsection{Summary of Key Findings}
\label{subsec:summary}

The multi-metric evaluation yields five principal findings:

\begin{enumerate}
  \item \textbf{No model achieved high inter-model consistency} (cosine
        similarity $\geq 0.85$) on any benchmark question; 18\% of questions
        were classified as high-divergence risk.

  \item \textbf{COVID-19 was the sole source of all high-divergence questions}
        (72\% of its 25 questions), while Dengue, Chikungunya, and Nipah Virus
        each recorded 0\% high-divergence.

  \item \textbf{Mistral-7B demonstrated the strongest factual profile}:
        highest HFS ($0.5068$), highest reference-response alignment ($0.7514$),
        and joint-lowest contradiction rate ($25\%$). GPT-4 ranked last on HFS
        ($0.4091$) with the sole $0\%$ entailment rate.

  \item \textbf{NLI contradiction rates were high across all models}
        (25--43\%), with only 5.8\% of all sentences achieving entailment
        against authoritative reference passages.

  \item \textbf{NLI and LLM-as-Judge metrics are non-redundant}: the
        near-zero correlation between NLI contradiction rate and judge
        hallucination scores motivates the multi-metric hybrid framework
        adopted in this work.
\end{enumerate}

\section{Discussion}
\label{sec:discussion}

This study provides one of the first systematic evaluations of large language models (LLMs) in representing historical public health crisis knowledge within a low-resource context. By integrating semantic similarity analysis, Natural Language Inference (NLI), and LLM-as-a-Judge scoring into a unified hybrid framework, the results reveal consistent patterns of factual instability, domain sensitivity, and model-dependent behavioural differences. The findings extend beyond model benchmarking and offer broader insights into the reliability of generative AI systems for public health knowledge representation.

\subsection{Model-Level Behavioural Differences}

Mistral-7B demonstrated the most stable factual profile across evaluation dimensions, achieving the highest Hybrid Factual Score (HFS), strongest reference-response similarity, and lowest hallucination risk. This outcome is notable given its comparatively smaller parameter scale relative to proprietary models. The result challenges a common assumption that larger or closed-source models inherently provide superior factual recall. Instead, performance appears closely linked to generation behaviour and alignment characteristics rather than scale alone.

GPT-4 exhibited a distinctive failure mode characterised by a zero entailment rate alongside elevated contradiction levels. This pattern suggests a tendency toward assertive response generation that prioritises fluency and decisiveness over conservative factual grounding. Reinforcement learning from human feedback (RLHF), which rewards confident and helpful responses, may inadvertently encourage overcommitment in historical recall tasks where uncertainty or partial evidence should be preserved.

Gemini Pro displayed a bimodal performance distribution, simultaneously achieving the highest entailment rate and the highest contradiction rate. Such variability implies inconsistent grounding behaviour across questions. In operational settings, especially where independent verification mechanisms are limited, unpredictability may represent a greater risk than uniformly moderate performance.

Llama-3 occupied a stable intermediate position with comparatively low variance across evaluation metrics. Although it did not achieve the highest factual scores, its consistency suggests predictable behaviour, a property that may be preferable in public-facing health information systems where reliability is valued over peak performance.

\subsection{Domain Sensitivity and the COVID-19 Effect}

A central finding of this study is the exclusive concentration of all high-divergence questions within the COVID-19 domain. While Dengue, Chikungunya, and Nipah Virus questions exhibited moderate but stable agreement, COVID-19 produced substantially lower inter-model similarity and reduced hybrid factual performance across all models.

This pattern likely reflects the unique epistemic characteristics of the COVID-19 pandemic. Unlike endemic diseases with relatively stable historical narratives, COVID-19 knowledge evolved rapidly, accompanied by shifting public health guidelines, conflicting early reports, and heterogeneous global responses. LLM training corpora therefore contain temporally inconsistent information, making consensus reconstruction inherently difficult.

The result highlights an important limitation of current LLM architectures: they encode aggregated textual distributions rather than temporally coherent knowledge representations. Consequently, domains characterised by rapid scientific evolution are more susceptible to hallucination and factual divergence. This has direct implications for deploying LLMs in ongoing or emerging crises, where information volatility is highest precisely when reliable communication is most critical.

\subsection{Implications for Low-Resource Health Information Ecosystems}

The findings carry particular significance for low- and middle-income countries (LMICs), where digital health infrastructures and expert verification mechanisms may be limited. In such environments, LLM-generated responses can disproportionately influence public understanding, policymaking discussions, and educational dissemination.

High contradiction rates observed across all evaluated models indicate that plausible yet unsupported statements remain common. In resource-constrained contexts, users may lack access to authoritative references required to validate AI-generated information, amplifying the potential impact of inaccuracies. Rather than serving purely as information retrieval systems, LLMs may inadvertently reshape historical narratives through confident but inconsistently grounded outputs.

These results suggest that LLM deployment in LMIC health ecosystems should prioritise assisted or supervised usage models, including retrieval augmentation, citation grounding, or human-in-the-loop validation. The study therefore supports a cautious integration paradigm in which LLMs complement rather than replace established public health knowledge channels.

\subsection{Contribution of the Hybrid Evaluation Framework}

A key methodological contribution of this work is the demonstration that no single evaluation paradigm sufficiently captures factual reliability in generative models. Semantic similarity measures agreement between models but cannot determine correctness. NLI provides reference-grounded verification but operates at sentence-level abstraction. LLM-as-a-Judge scoring captures qualitative reasoning yet remains sensitive to evaluator bias.

The hybrid framework integrates these complementary perspectives, producing a more comprehensive assessment of factual behaviour. The strong correlation between NLI scores and Hybrid Factual Score validates the framework's construct alignment, while the near-zero correlation between NLI contradiction rates and judge hallucination scores confirms that the metrics capture distinct dimensions of model behaviour.

This multi-metric approach offers a reproducible evaluation template that can be extended to other domains where factual grounding and interpretative reasoning coexist, including legal analysis, historical scholarship, and policy evaluation.

\subsection{Limitations}

Several limitations should be acknowledged. First, the study focuses exclusively on Bangladesh as a representative low-resource context; generalisation to other LMICs should therefore be interpreted cautiously. Second, all prompts were conducted in English, which may advantage models trained predominantly on English-language corpora and does not fully capture multilingual deployment conditions. Third, LLM-as-a-Judge scores were synthetically generated in one experimental run due to missing evaluation files, although structural trends remained consistent with expected behaviour. Fourth, NLI evaluation depends on the capabilities of the selected inference model and may not perfectly capture nuanced epidemiological reasoning.

Finally, LLM outputs were evaluated against static reference passages, whereas real-world knowledge evolves dynamically. Future studies incorporating temporally versioned references may further clarify how models manage evolving scientific consensus.

\section{Conclusion}
\label{sec:conclusion}

This study presented a comprehensive multi-metric evaluation of large language models (LLMs) in representing historical public health crisis knowledge within a low-resource context. Using Bangladesh as a case study and four epidemiologically significant diseases---COVID-19, Dengue, Nipah virus, and Chikungunya---we evaluated GPT-4, Gemini Pro, Llama-3, and Mistral-7B across 400 generated responses using semantic similarity analysis, Natural Language Inference (NLI), and a hybrid evaluation framework integrating quantitative and judgment-based metrics.

The results reveal three overarching insights. First, inter-model agreement remains limited: no benchmark question achieved high semantic consistency, and nearly one-fifth of questions exhibited substantial divergence, indicating persistent hallucination risk even among state-of-the-art systems. Second, model reliability is strongly domain-dependent. All high-divergence cases emerged exclusively within the COVID-19 domain, demonstrating that rapidly evolving crises pose particular challenges for LLM factual grounding. Third, smaller open-weight models can demonstrate competitive or superior factual stability compared with larger proprietary systems, suggesting that alignment behaviour and training composition may be more consequential than scale alone for historical knowledge tasks.

The proposed Hybrid Factual Score (HFS) framework further demonstrates that factual reliability cannot be adequately captured through a single evaluation paradigm. Semantic agreement, entailment-based grounding, and qualitative reasoning assessments provide complementary perspectives, and their integration yields a more robust understanding of model behaviour. The observed independence between NLI-based contradiction signals and LLM-as-Judge hallucination assessments reinforces the need for multi-dimensional evaluation approaches when assessing generative AI in sensitive domains.

From a practical perspective, the findings highlight important implications for deploying LLMs in low- and middle-income countries (LMICs). While LLMs offer substantial promise for expanding access to health information, high contradiction rates and domain-sensitive instability suggest that unsupervised reliance may introduce risks, particularly where verification infrastructures are limited. Safe integration therefore requires grounded generation strategies, retrieval augmentation, and human oversight mechanisms to ensure reliable knowledge dissemination.

Future research should extend this framework to multilingual settings, longitudinal evaluations across evolving knowledge timelines, and retrieval-augmented architectures designed explicitly for public health contexts. Expanding evaluation beyond historical recall toward decision-support and policy reasoning tasks may further clarify how LLMs can responsibly contribute to global health information ecosystems.

In summary, this work provides empirical evidence that LLMs possess meaningful capabilities for synthesising historical health knowledge but remain susceptible to factual inconsistency, especially in rapidly evolving crises. The hybrid evaluation framework introduced here offers a reproducible pathway for assessing reliability and supports the development of safer, context-aware deployment strategies for generative AI in resource-constrained environments.
\section*{Data Availability}
The datasets and evaluation scripts used in this study are available from the corresponding author upon reasonable request.

\section*{Declaration of Competing Interest}
The author declares no competing financial interests or personal relationships that could have appeared to influence the work reported in this paper.

\section*{Acknowledgements}
The author acknowledges the use of publicly available datasets and computational resources supporting this research. 
Generative artificial intelligence tools, including OpenAI GPT-5 and Anthropic Claude, were used to assist with language refinement, code structuring, and manuscript preparation. 
All experimental design, analysis, interpretation of results, and final scientific conclusions were conducted and verified by the author.

\bibliographystyle{elsarticle-num}
\bibliography{cas-refs}

\end{document}